\documentclass{article} % For LaTeX2e
\usepackage{microtype}
\usepackage{graphicx}
\usepackage{booktabs} % for professional tables

\usepackage{amsmath,amsfonts,bm}

\def\eqref#1{equation~\ref{#1}}
\def\1{\bm{1}}

\DeclareMathAlphabet{\mathsfit}{\encodingdefault}{\sfdefault}{m}{sl}
\SetMathAlphabet{\mathsfit}{bold}{\encodingdefault}{\sfdefault}{bx}{n}

\usepackage{hyperref}
\usepackage{url}
\usepackage{graphicx}
\usepackage{color}
\usepackage{subcaption}
\usepackage{algorithm}
\usepackage{algorithmic}

\renewcommand{\algorithmiccomment}[1]{\bgroup\hfill//~#1\egroup}

\usepackage[accepted]{icml2019}

\begin{document}

\twocolumn[
\icmltitle{Meta-Learning Neural Bloom Filters}
\begin{icmlauthorlist}
\icmlauthor{Jack W Rae}{dm,ucl}
\icmlauthor{Sergey Bartunov}{dm}
\icmlauthor{Timothy P Lillicrap}{dm,ucl}
\end{icmlauthorlist}

\icmlaffiliation{dm}{DeepMind, London, UK}
\icmlaffiliation{ucl}{CoMPLEX, Computer Science, University College London, London, UK}
\icmlcorrespondingauthor{Jack W Rae}{jwrae@google.com}
\icmlkeywords{Meta-Learning, Bloom Filter, Memory, Memory-Augmented Neural Networks, Compression, One-shot Learning}
\vskip 0.3in
]

\printAffiliationsAndNotice{}

\begin{abstract}
There has been a recent trend in training neural networks to replace data structures that have been crafted by hand, with an aim for faster execution, better accuracy, or greater compression. 
In this setting, a neural data structure is instantiated by training a network over many epochs of its inputs until convergence. In applications where inputs arrive at high throughput, or are ephemeral, training a network from scratch is not practical. This motivates the need for few-shot neural data structures.
In this paper we explore the learning of approximate set membership over a set of data in one-shot via meta-learning.
We propose a novel memory architecture, the Neural Bloom Filter, which is able to achieve significant compression gains over classical Bloom Filters and existing memory-augmented neural networks.
\end{abstract}

\section{Introduction}
One of the simplest questions one can ask of a set of data is whether or not a given query is contained within it. Is $q$, our query, a member of $S$, our chosen set of observations? This \textit{set membership} query arises across many computing domains; from databases, network routing, and firewalls. One could query set membership by storing $S$ in its entirety and comparing $q$ against each element. However, more space-efficient solutions exist.

The original and most widely implemented \textit{approximate set membership} data-structure is the Bloom Filter \citep{bloom1970space}. It works by storing sparse distributed codes, produced from randomized hash functions, within a binary vector. The Bloom-filter trades off space for an allowed false positive rate, which arises due to hash collisions. However its error is one-sided; if an element $q$ is contained in $S$ then it will always be recognized. It never emits false negatives.  One can find Bloom Filters embedded within a wide range of production systems; from \textit{network security} \citep{geravand2013bloom}, to block malicious IP addresses; \textit{databases}, such as Google's Bigtable \citep{chang2008bigtable}, to avoid unnecessary disk lookups; \textit{cryptocurrency} \citep{hearn2012connectio}, to allow clients to filter irrelevant transactions; \textit{search}, such as Facebook's typeahead search \citep{adams2010life}, to filter pages which do not contain query prefixes; and \textit{program verification} \citep{dillinger2004bloom}, to avoid recomputation over previously observed states. 

While the main appeal of Bloom Filters is favourable compression, another important quality is the support for dynamic updates. New elements can be inserted in $\mathcal{O}(1)$ time. This is not the case for all approximate set membership data structures. For example, perfect hashing saves $\approx 40\%$ space over Bloom Filters but requires a pre-processing stage that is polynomial-time in the number of elements to store \citep{dietzfelbinger2008succinct}. Whilst the static set membership problem is interesting, it limits the applicability of the algorithm. For example, in a database application that is serving a high throughput of write operations, it may be intractable to regenerate the full data-structure upon each batch of writes. 

We thus focus on the data stream computation model \citep{muthukrishnan2005data}, where input observations are assumed to be ephemeral and can only be inspected a constant number of times --- usually once. This captures many real-world applications: network traffic analysis, database query serving, and reinforcement learning in complex domains. Devising an approximate set membership data structure that is not only more compressive than Bloom Filters, but can be applied to either dynamic or static sets, could have a significant performance impact on modern computing applications. In this paper we investigate this problem using memory-augmented neural networks and meta-learning.

We build upon the recently growing literature on using neural networks to replace algorithms that are configured by heuristics, or do not take advantage of the data distribution. For example, Bloom Filters are indifferent to the data distribution. They have near-optimal space efficiency when data is drawn uniformly from a universe set \citep{carter1978exact} (maximal-entropy case) but (as we shall show) are sub-optimal when there is more structure. Prior studies on this theme have investigated compiler optimization \citep{cummins2017end}, computation graph placement \citep{azalia2017device}, and data index structures such as b-trees \citep{kraska2018case}. In the latter work, \citet{kraska2018case} explicitly consider the problem of static set membership. By training a neural network over a fixed $S$ (in their case, string inputs) along with held-out negative examples, they observe $36\%$ space reduction over a conventional Bloom Filter\footnote{The space saving increases to $41\%$ when an additional trick is incorporated, in discretizing and re-scaling the classifier outputs and treating the resulting function as a hash function to a bit-map.}. Crucially this requires iterating over the storage set $S$ a large number of times to embed its salient information into the weights of a neural network classifier. For a new $S$ this process would have to be repeated from scratch.

Instead of learning from scratch, we draw inspiration from the few-shot learning advances obtained by meta-learning memory-augmented neural networks \citep{santoro2016meta, vinyals2016matching}. In this setup, tasks are sampled from a common distribution and a network learns to specialize to (learn) a given task with few examples. This matches very well to applications where many Bloom Filters are instantiated over different subsets of a common data distribution. For example, a Bigtable database usually contains one Bloom Filter per SSTable file. For a large table that contains Petabytes of data, say, there can be over $100,000$ separate instantiated data-structures which share a common row-key format and query distribution. Meta-learning allows us to exploit this common redundancy. We design a database task with similar redundancy to investigate this exact application in Section \ref{sec:database_queries}.

The main contributions of this paper are (1) A new memory-augmented neural network architecture, the \textit{Neural Bloom Filter}, which learns to write to memory using a distributed write scheme, and (2) An empirical evaluation of the Neural Bloom Filter meta-learned on one-shot approximate set membership problems of varying structure. We compare with the classical Bloom Filter alongside other memory-augmented neural networks such as the Differentiable Neural Computer \citep{graves2016hybrid} and Memory Networks \citep{sukhbaatar2015end}. We find when there is no structure, that differentiates the query set elements and queries, the Neural Bloom Filter learns a solution similar to a Bloom Filter derivative --- a Bloom-g filter \citep{qiao2011one} --- but when there is a lot of structure the solution can be considerably more compressive (e.g. $30\times$ for a database task).
\section{Background}
\subsection{Approximate Set Membership}
The problem of \textit{exact set membership} is to state whether or not a given query $q$ belongs to a set of $n$ distinct observations $S = \{x_1, \ldots, x_n \}$ where $x_i$ are drawn from a universe set $U$. By counting the number of distinct subsets of size $n$ it can be shown that any such exact set membership tester requires at least $\log_2 \binom{|U|}{n}$ bits of space. To mitigate the space dependency on $|U|$, which can be prohibitively large, one can relax the constraint on perfect correctness. \textit{Approximate set membership} allows for a false positive rate of at most $\epsilon$. Specifically we answer $q \in A(S)$ where $A(S) \supseteq S$ and $p(q \in A(S) - S) \le \epsilon$. It can be shown\footnote{By counting the minimal number of $A(S)$ sets required to cover all $S \subset U$.} the space requirement for approximate set membership of uniformly sampled observations is at least $n\log_2(\frac{1}{\epsilon})$ bits \citep{carter1978exact} which can be achieved with perfect hashing. So for a false positive rate of $1\%$, say, this amounts to $6.6$ bits per element. In contrast to storing raw or compressed elements this can be a huge space saving, for example ImageNet images require $108$ KB per image on average when compressed with JPEG, an increase of over four orders of magnitude. 
\subsection{Bloom Filter}
The Bloom Filter \citep{bloom1970space} is a data structure which solves the dynamic approximate set membership problem with near-optimal space complexity. It assumes access to k uniform hash functions $h_i : U \rightarrow \{1, \ldots, m \}, \; i = 1, \ldots, k$ such that $p(h_i(x) = j) = 1/m$ independent of prior hash values or input $x$. The Bloom Filter's memory $M \in [0, 1]^m$ is a binary string of length $m$ which is initialized to zero. Writes are performed by hashing an input $x$ to $k$ locations in $M$ and setting the corresponding bits to $1$, $M[h_i(x)] \leftarrow 1; \; i = 1, \ldots, k$. For a given query $q$ the Bloom Filter returns true if all corresponding hashed locations are set to $1$ and returns false otherwise: $Query(M, q):= M[h_1(q)] \land M[h_2(q)] \land \ldots \land M[h_k(q)]$. This incurs zero false negatives, as any previously observed input must have enabled the corresponding bits in $M$, however there can be false positives due to hash collisions. To achieve a false positive rate of $\epsilon$ with minimal space one can set $k = \log_2{(1 / \epsilon)}$ and $m = n \log_2{(1 / \epsilon)} \log_2{e}$, where $e$ is Euler's number. The resulting space is a factor of $\log_2{e} \approx 1.44$ from the optimal static lower bound given by \citet{carter1978exact}.
\subsection{Memory-Augmented Neural Networks}
Recurrent neural networks such as LSTMs retain a small amount of memory via the recurrent state. However this is usually tied to the number of trainable parameters in the model. There has been recent interest in augmenting neural networks with a larger external memory. The method for doing so, via a differentiable write and read interface, was first popularized by the Neural Turing Machine (NTM) \citep{graves2014neural} and its successor the Differentiable Neural Computer (DNC) \citep{graves2016hybrid} in the context of learning algorithms, and by Memory Networks \citep{sukhbaatar2015end} in the context of question answering. Memory Networks store embeddings of the input in separate rows of a memory matrix $M$. Reads are performed via a differentiable \textit{content-based addressing} operation. Given a query embedding $q$ we take some similarity measure $D$ (e.g. cosine similarity, or negative euclidean distance) against each row in memory and apply a softmax to obtain a \textit{soft} address vector $a \propto e^{D(q, M)}$. A read is then a weighted sum over memory $r \leftarrow a^TM$. The NTM and DNC use the same content-based read mechanism, but also learns to write. These models can arbitrate whether to write to slots in memory with similar content (content-based writes), temporally ordered locations, or unused memory.

When it comes to capacity, there has been consideration to scaling both the DNC and Memory Networks to very large sizes using sparse read and write operations \citep{rae2016scaling, chandar2016hierarchical}. However another way to increase the capacity is to increase the amount of compression which occurs in memory. Memory Nets can create compressive representations of each input, but cannot compress jointly over multiple inputs because they are hard-wired to write one slot per timestep. The NTM and DNC can compress over multiple slots in memory because they can arbitrate writes across multiple locations, but in practice seem to choose very sharp read and write addresses. The Kanerva Machine \citep{wu201kanerva, wu2018learning} tackles memory-wide compression using a distributed write scheme to jointly compose and compress its memory contents. The model uses content-based addressing over a separate learnable addressing matrix $A$, instead of the memory $M$, and thus learns \textit{where} to write. We take inspiration from this scheme.
\section{Model}
%\textbf{Model} 
\label{sec:model}
\vspace{-6px}
\begin{figure*}
    \centering
    \includegraphics[width=0.8\textwidth]{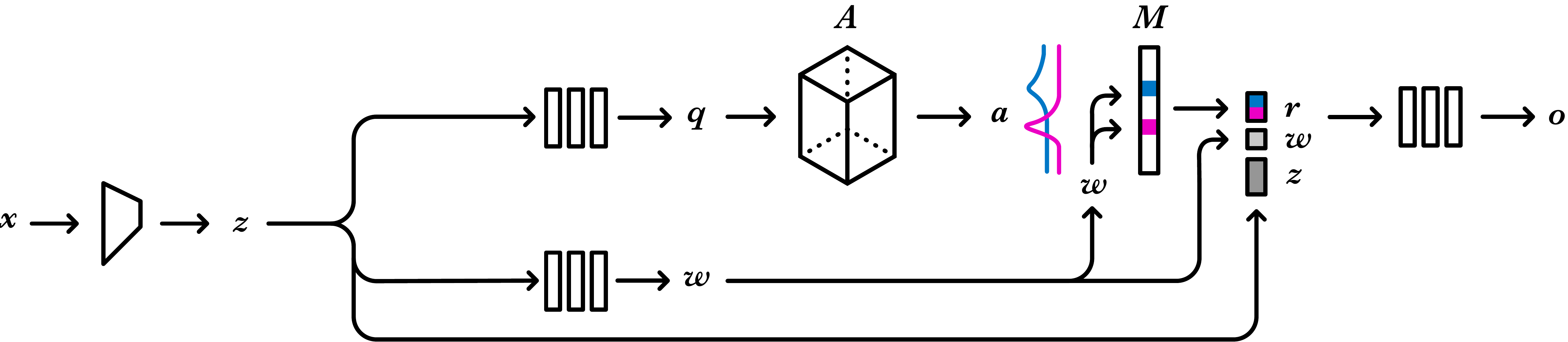}
    \caption{Overview of the Neural Bloom Filter architecture.}
    \label{fig:model_diagram}
\end{figure*}
One approach to learning set membership in one-shot would be to use a recurrent neural network, such as an LSTM or DNC. Here, the model sequentially ingests the $N$ elements to store, answers a set of queries using the final state, and is trained by BPTT. Whilst this is a general training approach, and the model may learn a compressive solution, it does not scale well to larger number of elements. Even when $N = 1000$, backpropagating over a sequence of this length induces computational and optimization challenges. For larger values this quickly becomes intractable.  Alternatively one could store an embedding of each element $x_i \in S$ in a slot-based Memory Network. This is more scalable as it avoids BPTT, because the gradients of each input can be calculated in parallel. However Memory Networks are not a space efficient solution (as shown in in Section \ref{sec:experiments}) because there is no joint compression of inputs.
\begin{algorithm}[h]
    \caption{Neural Bloom Filter}
    \label{alg:nbf}
    \begin{algorithmic}[1]
    \STATE def controller(x):
    \STATE $\quad z \leftarrow f_{enc}(x)$ \COMMENT{Input embedding} \\
    \STATE $\quad q \leftarrow f_q(z)$ \COMMENT{Query word} \\
    \STATE $\quad a \leftarrow \sigma(q^T A)$ \COMMENT{Memory address} \\
    \STATE $\quad w \leftarrow f_{w}(z) $ \COMMENT{Write word} \\ 
    \item[]
    \STATE def write(x):
    \STATE $\quad a, w \leftarrow \hbox{controller}(x)$
    \STATE $\quad M_{t + 1} \leftarrow M_t + w a^T$ \COMMENT{Additive write} \\
    \item[]
    \STATE def read(x):
    \STATE $\quad a, w, z \leftarrow \hbox{controller}(x)$
    \STATE $\quad r \leftarrow \hbox{flatten}(M \odot$ a) \COMMENT{Read words} \\
    \STATE $\quad o \leftarrow f_{out}([r, w, z])$ \COMMENT{Output logit} \\
    \end{algorithmic}
\end{algorithm}

This motivates the proposed memory model, the Neural Bloom Filter. Briefly, the network is augmented with a real-valued memory matrix. The network \textit{addresses} memory by classifying which memory slots to read or write to via a softmax, conditioned on the input. We can think of this as a continuous analogue to the Bloom Filter's hash function; because it is learned the network can co-locate or separate inputs to improve performance. The network updates memory with a simple \textit{additive write} operation --- i.e. no multiplicative gating or squashing --- to the addressed locations. An additive write operation can be seen as a continuous analogue to the the Bloom Filter's logical OR write operation. Crucially, the additive write scheme allows us to train the model without BPTT --- this is because gradients with respect to the write words $\partial L / \partial w = (\partial L / \partial M)^T a$ can be computed in parallel. Reads involve a component-wise multiplication of address and memory (analogous to the selection of locations in the Bloom Filter via hashing), but instead of projecting this down to a scalar with a fixed function, we pass this through an MLP to obtain a scalar familiarity logit. The network is fully differentiable, allows for memories to be stored in a distributed fashion across slots, and is quite simple e.g. in comparison to DNCs.

The full architecture depicted in Figure \ref{fig:model_diagram} consists of a \textit{controller network} which encodes the input to an embedding $z \leftarrow f_{enc}(x)$ and transforms this to a write word $w \leftarrow f_w(z)$ and a query $q \leftarrow f_q(z)$. The address over memory is computed via a softmax $a \leftarrow \sigma(q^T A)$ over the content-based attention between $q$ and a learnable address matrix $A$. Here, $\sigma$ denotes a softmax. The network thus learns where to place elements or overlap elements based on their content, we can think of this as a soft and differentiable relaxation of the uniform hashing families incorporated by the Bloom Filter (see Appendix \ref{app:hashing_relation} for further discussion).

A \textit{write} is performed by running the controller to obtain a write word $w$ and address $a$, and then additively writing $w$ to $M$, weighted by the address $a$, $M_{t + 1} \leftarrow M_t + w a^T$. The simple additive write ensures that the resulting memory is invariant to input ordering (as addition is commutative) and we do not have to backpropagate-through-time (BPTT) over sequential writes --- gradients can be computed in parallel.  A \textit{read} is performed by also running the controller network to obtain $z, w,$ and $a$ and component-wise multiplying the address $a$ with $M$, $r \leftarrow M \odot a$. The read words $r$ are fed through an MLP along with the residual inputs $w$ and $z$ and are projected to a single scalar logit, indicating the familiarity signal. We found this to be more powerful than the conventional read operation $r \leftarrow a^T M$ used by the DNC and Memory Networks, as it allows for non-linear interactions between rows in memory at the time of read. See Algorithm \ref{alg:nbf} for an overview of the operations.

To give an example network configuration, we chose $f_{enc}$ to be a 3-layer CNN in the case of image inputs, and a $128$-hidden-unit LSTM in the case of text inputs. We chose $f_w$ and $f_q$ to be an MLP with a single hidden layer of size $128$, followed by layer normalization, and $f_{out}$ to be a 3-layer MLP with residual connections. We used a leaky ReLU as the non-linearity. Although the described model uses dense operations that scale linearly with the memory size $m$, we discuss how the model could be implemented for $\mathcal{O}(\log m)$ time reads and writes using sparse attention and read/write operations, in Appendix \ref{app:efficient_addressing}. Furthermore the model's relation to uniform hashing is discussed in Appendix \ref{app:hashing_relation}. 
\section{Space Complexity}
In this section we discuss space lower bounds for the approximate set membership problem when there is some structure to the storage or query set. This can help us formalise why and where neural networks may be able to beat classical lower bounds to this problem.

The $n\log_2{(1 / \epsilon)}$ lower bound from \citet{carter1978exact} assumes that all subsets $S \subset U$ of size $n$, and all queries $q \in U$ have equal probability. Whilst it is instructive to bound this maximum-entropy scenario, which we can think of as `worst case', most applications of approximate set membership  e.g. web cache sharing, querying databases, or spell-checking, involve sets and queries that are not sampled uniformly. For example, the elements within a given set may be highly dependent, there may be a power-law distribution over queries, or the queries and sets themselves may not be sampled independently.

A more general space lower bound can be defined by an information theoretic argument from communication complexity \citep{yao1979some}. Namely, approximate set membership can be framed as a two-party communication problem between Alice, who observes the set $S$ and Bob, who observes a query $q$. They can agree on a shared policy $\Pi$ in which to communicate. For given inputs $S, q$ they can produce a transcript $A_{S, q} = \Pi(S, q) \in \mathcal{Z}$ which can be processed $g : \mathcal{Z} \rightarrow {0, 1}$ such that $\mathbb{P}\left(g(A_{S, q}) = 1 | q \not\in S \right) \le \epsilon$. \citet{bar2004information} shows that the maximum transcript size is greater than the mutual information between the inputs and transcript: $\max_{S, q} |A_{S, q}| \ge I\!\left(S, q ; A_{S, q}\right) = H(S, q) - H(S, q | A_{S, q})$. Thus we note problems where we may be able to use less space than the classical lower bound are cases where the entropy $H(S, q)$ is small, e.g. our sets are highly non-uniform, or cases where $H(S, q | A_{S, q})$ is large, which signifies that many query and set pairs can be solved with the same transcript.  

\section{Experiments}
\label{sec:images}
\begin{figure*}[h]
    \centering
    \includegraphics[width=0.8\textwidth]{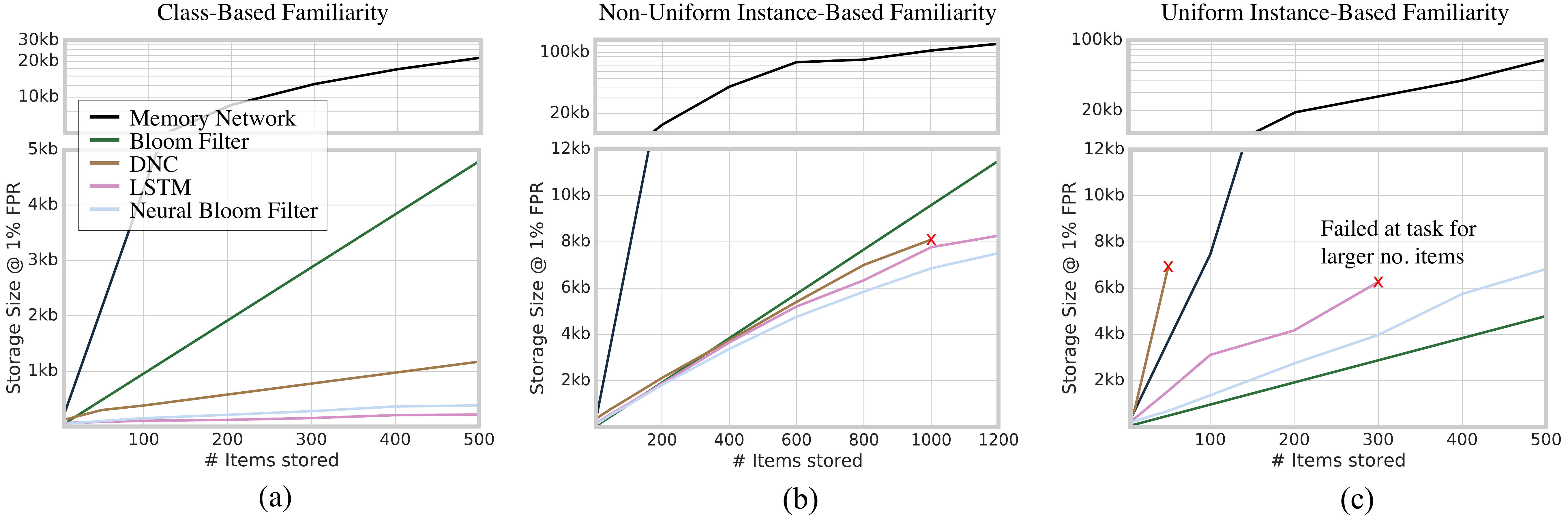}
    \caption{\textbf{Sampling strategies on MNIST.} Space consumption at 1\% FPR.}
    \label{fig:mnist_tasks}
\end{figure*}
\label{sec:experiments}

Our experiments explore scenarios where set membership can be learned in one-shot with improved compression over the classical Bloom Filter. We consider tasks with varying levels of structure in the storage sets $S$ and queries $q$. We compare the Neural Bloom Filter with three memory-augmented neural networks, the LSTM, DNC, and Memory Network, that are all able to write storage sets in one-shot. 

\begin{algorithm}[tb]
    \caption{Meta-Learning Training}
    \label{alg:training}
    \begin{algorithmic}[1]
    \STATE Let $S^{train}$ denote the distribution over sets to store.
    \STATE Let $Q^{train}$ denote the distribution over queries.
    \FOR{$i = 1$ {\bfseries to} max train steps}
    \STATE Sample task:
    \STATE $\quad$ Sample set to store: $S \sim \mathcal{S}^{train}$ 
    \STATE $\quad$ Sample $t$ queries: $x_1, \ldots, x_t \sim Q^{train}$
    \STATE $\quad$ Targets: $y_j = 1 \hbox{ if } x_j \in S \hbox{ else } 0; \; j = 1, \ldots, t$
    \STATE Write entries to memory: $M \leftarrow f_{\theta}^{write}(S)$
    \STATE Calculate logits: $o_j = f_{\theta}^{read}(M, x_j); \; j = 1, \ldots, t$
    \STATE XE loss: $L = \sum_{j = 1}^t y_j \log{o_j} + (1 - y_j)(1 - \log{o_j})$ 
    \STATE Backprop through queries and writes: $dL / d\theta$
    \STATE Update parameters: $\theta_{i + 1} \leftarrow \hbox{Optimizer}(\theta_i, dL / d\theta)$
    \ENDFOR
    \end{algorithmic}
\end{algorithm}
The training setup follows the memory-augmented meta-learning training scheme of \citet{vinyals2016matching}, only here the task is familiarity classification versus image classification. The network samples tasks which involve classifying  familiarity for a given storage set. Meta-learning occurs as a two-speed process, where the model quickly learns to recognize a given storage set $S$ within a training episode via writing to a memory or state, and the model slowly learns to improve this fast-learning process by optimizing the model parameters $\theta$ over multiple tasks. We detail the training routine in Algorithm \ref{alg:training}.

For the RNN baselines (LSTM and DNC) the write operation corresponds to unrolling the network over the inputs and outputting the final state. For these models, the query network is simply an MLP classifier which receives the concatenated final state and  query, and outputs a scalar logit. For the Memory Network, inputs are stored in individual slots and the familiarity signal is computed from the maximum content-based attention value. The Neural Bloom Filter read and write operations are defined in Algorithm \ref{alg:nbf}.
\subsection{Space Comparison}
We compared the space (in bits) of the model's memory (or state) to a Bloom Filter at a given false positive rate and $0\%$ false negative rate. The false positive rate is measured empirically over a sample of $50,000$ queries for the learned models; for the Bloom Filter we employ the analytical false positive rate. Beating a Bloom Filter's space usage with the analytical false positive rate implies better performance for any given Bloom Filter library version (as actual Bloom Filter hash functions are not uniform), thus the comparison is reasonable. For each model we sweep over hyper-parameters relating to model size to obtain their smallest operating size at the desired false positive rate (for the full set, see Appendix \ref{app:hyperparams}). Because the neural models can emit false negatives, we store these in a (ideally small) backup Bloom Filter, as proposed by \citet{kraska2018case, mitzenmacher2018model}. We account for the space of this backup Bloom Filter, and add it to the space usage of the model's memory for parity (See Appendix \ref{app:space_comparison} for further discussion). The neural network must learn to output a small state in one-shot that can serve set membership queries at a given false positive rate, and emit a small enough number of false negatives such that the backup filter is also small, and the total size is considerably less than a Bloom Filter.
\subsection{Sampling Strategies on MNIST}
\begin{figure*}[h]
    \centering
    \includegraphics[width=0.85\textwidth]{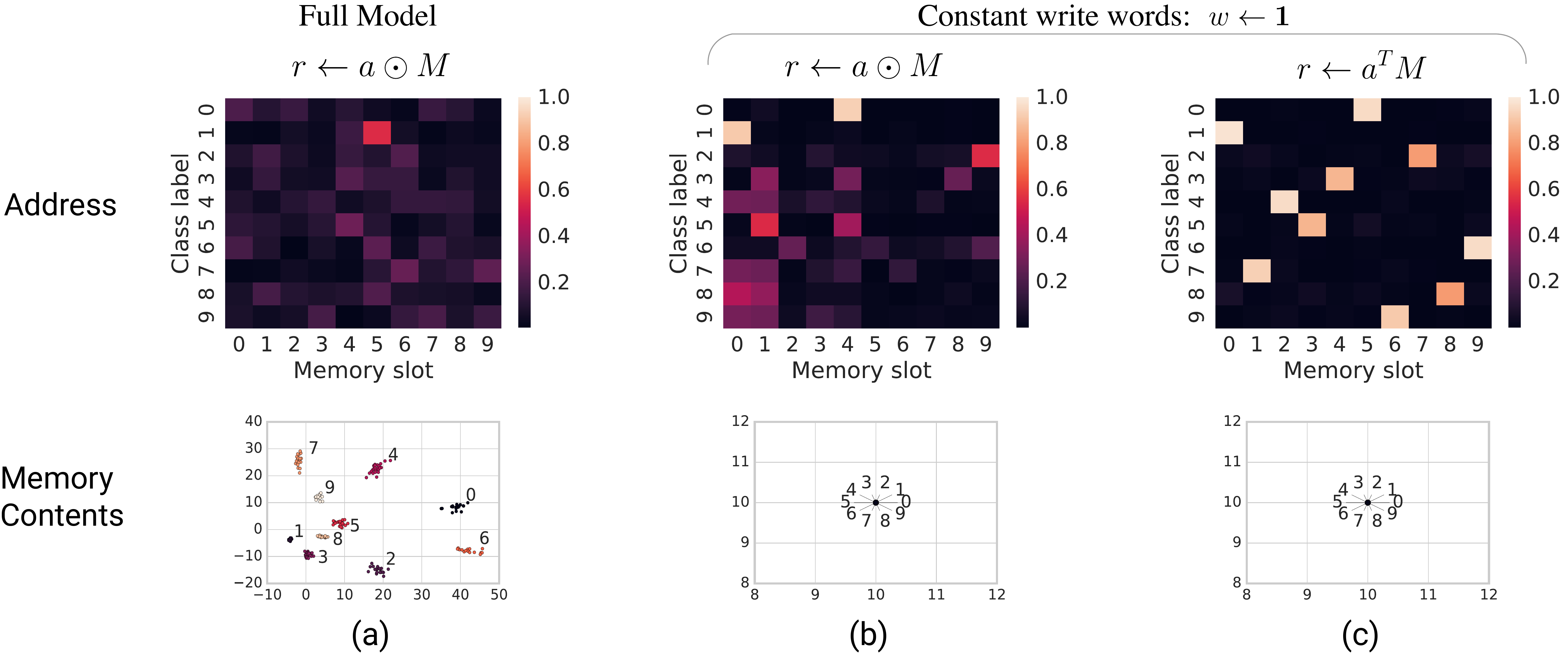}
    \caption{\textbf{Memory access analysis.} Three different learned solutions to class-based familiarity. We train three Neural Bloom Filter variants, with a succession of simplified read and write mechanisms. Each model contains $10$ memory slots and the memory addressing weights $a$ and contents $\bar{M}$ are visualised, broken down by class. Solutions share broad correspondence to known algorithms: (a) Bloom-g filters, (b) Bloom Filters, (c) Perfect hashing.}
    \label{fig:mnist_analysis}
\end{figure*}
To understand what kinds of scenarios neural networks may be more (or less) compressive than classical Bloom Filters, we consider three simple set membership tasks that have a graded level of structure to the storage sets and queries. Concretely, they differ in the sampling distribution of storage sets $\mathcal{S}^{train}$ and queries $\mathcal{Q}^{train}$. However all problems are approximate set membership tasks that can be solved by a Bloom Filter. The tasks are \textit{(1) Class-based familiarity}, a highly structured task where each set of images is sampled with the constraint that they arise from the same randomly-selected class. \textit{(2) Non-uniform instance-based familiarity}, a moderately structured task where the images are sampled without replacement from an exponential distribution. \textit{(3) Uniform instance-based familiarity}, a completely unstructured task where each subset contains images sampled uniformly without replacement. For each task we varied the size of the sample set to store, and calculated the space (in bits) of each model's state at a fixed false positive rate of $1\%$ and a false negative rate of $0\%$. We used relatively small storage set sizes (e.g. $100 - 1000$) to start with, as this highlights that some RNN-based approaches struggle to train over larger set sizes, before progressing to larger sets in subsequent sections. See Appendix \ref{app:sampling} for further details on the task setup. In the \textit{class-based sampling} task we see in Figure \ref{fig:mnist_tasks}a that the DNC, LSTM and Neural Bloom Filter are able to significantly outperform the classical Bloom Filter when images are sampled by class. The Memory Network is able to solve the task with a word size of only $2$, however this corresponds to a far greater number of bits per element, $64$ versus the Bloom Filter's $9.8$ (to a total size of $4.8$kb), and so the overall size was prohibitive. The DNC, LSTM, and Neural Bloom Filter are able to solve the task with a storage set size of $500$ at $1.1$kb , $217$b, and $382$b; a $4.3\times$, $22\times$, and $12\times$ saving respectively.  For the \textit{non-uniform sampling task} in Figure \ref{fig:mnist_tasks}b we see the Bloom Filter is preferable for less than $500$ stored elements, but is overtaken thereafter. At $1000$ elements the DNC, LSTM, and Neural Bloom Filter consume $7.9$kb, $7.7$kb, and $6.8$kb respectively which corresponds to a $17.6\%$, $19.7\%$, and $28.6\%$ reduction over the $9.6$kb Bloom Filter. In the \textit{uniform sampling task} shown in Figure \ref{fig:mnist_tasks}c, there is no structure to the sampling of $S$. The two architectures which rely on BPTT essentially fail to solve the task at some threshold of storage size. The Neural Bloom Filter solves it with $6.8$kb (using a memory size of $50$ and word size of $2$). The overall conclusion from these sets of experiments is that the classical Bloom Filter works best when there is no structure to the data, however when there is (e.g. skewed data, or highly dependent sets that share common attributes) we do see significant space savings.
\subsection{Memory Access Analysis}
We wanted to understand how the Neural Bloom Filter uses its memory, and in particular how its learned solutions may correspond to classical algorithms. We inspected the memory contents (what was stored to memory) and addressing weights (where it was stored) for a small model of $10$ memory slots and a word size of $2$, trained on the MNIST class-based familiarity task. We plot this for each class label, and compare the pattern of memory usage to two other models that use increasingly simpler read and write operations: (1) an ablated model with constant write words $w \leftarrow \mathbf{1}$, and (2) an ablated model with $w \leftarrow \mathbf{1}$ \textit{and} a linear read operator $r \leftarrow a^T M$. 

The full model, shown in Figure \ref{fig:mnist_analysis}a learns to place some classes in particular slots, e.g. class $1 \rightarrow$ slot $5$, however most are distributed. Inspecting the memory contents, it is clear the write word encodes a unique $2$D token for each class. This solution bears resemblance with Bloom-g Filters \citep{qiao2011one} where elements are spread across a smaller memory with the same hashing scheme as Bloom Filters, but a unique token is stored in each slot instead of a constant $1$-bit value. With the model ablated to store only $1$s in Figure \ref{fig:mnist_analysis}b we see it uses semantic addressing codes for some classes (e.g. $0$ and $1$) and distributed addresses for other classes. E.g. for class $3$ the model prefers to uniformly spread its writes across memory slot $1$, $4$, and $8$. The model solution is similar to that of Bloom Filters, with distributed addressing codes as a solution --- but no information in the written words themselves. When we force the read operation to be linear in Figure \ref{fig:mnist_analysis}c, the network maps each input class to a unique slot in memory. This solution has a correspondence with perfect hashing. In conclusion, with small changes to the read/write operations we see the Neural Bloom Filter learn different algorithmic solutions.
\vspace{-0.8em}
\subsection{Database Queries}
\label{sec:database_queries}
\begin{table}[]
    \centering
    \begin{tabular}{r c c c}
    & 5\% & 1\% & 0.1\% \\ 
    Neural Bloom Filter &  871b  & 1.5kb  & 24.5kb \\
    Bloom Filter & 31.2kb  & 47.9kb  & 72.2kb  \\ 
    Cuckoo Filter & 33.1kb  &  45.3kb & 62.6kb   \\
    \end{tabular}
    \caption{\textbf{Database task}. Storing $5000$ row-key strings for a target false positive rate.} 
    \label{tab:database_large}
\end{table}
We look at a task inspired by database interactions. NoSQL databases, such as Bigtable and Cassandra, use a single string-valued row-key, which is used to index the data. The database is comprised of a union of files (e.g. SSTables) storing contiguous row-key chunks. Bloom Filters are used to determine whether a given query $q$ lies within the stored set. We emulate this setup by constructing a universe of strings, that is alphabetically ordered, and by sampling contiguous ranges (to represent a given SSTable). Queries are sampled uniformly from the universe set of strings. We choose the $2.5M$ unique tokens in the GigaWord v5 news corpus to be our universe as this consists of structured natural data and some noisy or irregular strings. 

We consider the task of storing sorted string sets of size $5000$. We train the Neural Bloom Filter to several desired false positive rates ($5\%, 1\%, 0.1\%$) and used a backup Bloom Filter to guarantee $0\%$ false negative rate. We also trained LSTMs and DNCs for comparison, but they failed to learn a solution to the task after several days of training; optimizing insertions via BPTT over a sequence of length $5000$ did not result in a remotely usable solution. The Neural Bloom Filter avoids BPTT via its simple additive write scheme, and so it learned to solve the task quite naturally. As such, we compare the Neural Bloom Filter solely to classical data structures: Bloom Filters and Cuckoo Filters. In Table \ref{tab:database_large} we see a significant space reduction of $3-40\times$, where the margin grows with increasing permitted false positive rates. Since memory is an expensive component within production databases (in contrast to disk, say), this memory space saving could translate to a non-trivial cost reduction. We note that a storage size of $5000$ may appear small, but is relevant to the NOSQL database scenario where disk files (e.g. SSTables) are typically sharded to be several megabytes in size, to avoid issues with compaction. E.g. if the stored values were of size $10$kB per row, we would expect $5000$ unique keys or less in an average Bigtable SSTable.

One further consideration for production deployment is the ability to extrapolate to larger storage set sizes during evaluation. We investigate this for the Neural Bloom Filter on the same database task, and compare it to an LSTM. To ensure both models train, we set the maximum training storage set size to $200$ and evaluate up to sizes $250$, a modest $25\%$ size increase. We find that the Neural Bloom Filter uses up to $3\times$ less space than the LSTM and the neural models are able to extrapolate to larger set sizes than those observed during training (see Appendix \ref{app:database_extrapolation} Figure \ref{fig:database_results}). Whilst the performance eventually degrades when the training limit size is exceeded, it is not catastrophic for either the LSTM or Neural Bloom Filter.

\subsection{Timing benchmark}
\begin{table*}[]
    \centering
    \begin{tabular}{c c c | c c | c c}
    %\hline
                                & \multicolumn{2}{c|}{Query + Insert Latency} & \multicolumn{2}{|c|}{Query Throughput (QPS)} & \multicolumn{2}{c}{Insert Throughput (IPS)} \\
                                & \small{CPU} & \small{GPU} & \small{CPU} & \small{GPU} & \small{CPU} & \small{GPU}       \\
                                %\hline
        Bloom Filter*                          & \textbf{0.02ms} & - &  61K & - & 61K & -   \\
        Neural Bloom Filter                    & 5.1ms & 13ms & 3.5K & 105K & 3.2K & \textbf{101K}  \\
        LSTM                                   & 5.0ms   & 13ms & 3.1K & \textbf{107K} & 2.4K & 4.6K \\
        Learned Index \citep{kraska2018case}   & 780ms   &  1.36s  & 3.1K  & \textbf{107K}  &  25   & 816 \\
        %\hline
    \end{tabular}
    \caption{Latency for a single query, and throughput for a batch of 10,000 queries. *Query-efficient Bloom Filter from~\citet{chen2007new}.}
    \label{tab:timing_benchmark}
\end{table*}
\label{app:timing}
We have principally focused on space comparisons in this paper, we now consider speed for the database task described in the prior section. We measure latency as the wall-clock time to complete a single insertion or query of a row-key string of length $64$. We also measure throughput as the reciprocal wall-clock time of inserting or querying $10,000$ strings. We use a common encoder architecture for the neural models, a 128-hidden-unit character LSTM. We benchmark the models on the CPU (Intel(R) Xeon(R) CPU E5-1650 v2 @ 3.50GHz) and on the GPU (NVIDIA Quadro P6000) with models implemented in TensorFlow without any model-specific optimizations. We compare to empirical timing results published in a query-optimized Bloom Filter variant \citep{chen2007new}. We include the Learned Index from \cite{kraska2018case} to contrast timings with a model that is not one-shot. The architecture is simply the LSTM character encoder; inserts are performed via gradient descent. The number of gradient-descent steps to obtain convergence is domain-dependent, we chose $50$ steps in our timing benchmarks. The Learned Index queries are obtained by running the character LSTM over the input and classifying familiarity --- and thus query metrics are identical to the LSTM baseline. 

We see in Table \ref{tab:timing_benchmark}. that the combined query and insert latency of the Neural Bloom Filter and LSTM sits at $5$ms on the CPU, around $400\times$ slower than the classical Bloom Filter. The Learned Index contains a much larger latency of $780$ms due to the sequential application of gradients. For all neural models, latency is not improved when operations are run on the GPU. However when multiple queries are received, the throughput of GPU-based neural models surpasses the classical Bloom Filter due to efficient concurrency of the dense linear algebra operations. This leads to the conclusion that a Neural Bloom Filter could be deployed in scenarios with high query load without a catastrophic decrease in throughput, if GPU devices are available. For insertions we see a bigger separation between the one-shot models: the LSTM and Neural Bloom Filter. Whilst all neural models are uncompetitive on the CPU, the Neural Bloom Filter surpasses the Bloom Filter's insertion throughput when placed on the GPU, with $101K$ insertions per second (IPS). The LSTM runs at $4.6K$ IPS, one order of magnitude slower, because writes are serial, and the Learned Index structure is two orders of magnitude slower at $816$ IPS due to sequential gradient computations. The benefits of the Neural Bloom Filter's simple write scheme are apparent here.

\section{Related Work}
There have been a large number of Bloom Filter variants published; from \textit{Counting Bloom Filters} which support deletions \citep{fan2000summary}, \textit{Bloomier Filters} which store functions vs sets \citep{chazelle2004bloomier}, \textit{Compressed Bloom Filters} which use arithmetic encoding to compress the storage set \citep{mitzenmacher2002compressed}, and \textit{Cuckoo Filters} which use cuckoo hashing to reduce redundancy within the storage vector \citep{fan2014cuckoo}. Although some of these variants focus on better compression, they do not achieve this by specializing to the data distribution. 

One of the few works which address data-dependence are \textit{Weighted Bloom Filters} \citep{bruck2006weighted, wang2015improved}. They work by modulating the number of hash functions used to store or query each input, dependent on its storage and query frequency. This requires estimating a large number of separate storage and query frequencies. This approach can be useful for imbalanced data distributions, such as the non-uniform instance-based MNIST familiarity task. However it cannot take advantage of dependent sets, such as the class-based MNIST familiarity task, or the database query task. We see the Neural Bloom Filter is more compressive in all settings.

\citet{sterne2012efficient} proposes a neurally-inspired set membership data-structure that works by replacing the randomized hash functions with a randomly-wired computation graph of \textit{OR} and \textit{AND} gates. The false positive rate is controlled analytically by modulating the number of gates and the overall memory size. However there is no learning or specialization to the data with this setup. \citet{bogacz2003comparison} investigates a learnable neural familiarity module, which serves as a biologically plausible model of familiarity mechanisms in the brain, namely within the perirhinal cortex. However this has not shown to be empirically effective at exact matching.

\citet{kraska2018case} consider the use of a neural network to classify the membership of queries to a fixed set $S$. Here the network itself is more akin to a perfect hashing setup where multiple epochs are required to find a succinct holistic representation of the set, which is embedded into the weights of the network. In their case this search is performed by gradient-based optimization. We emulate their experimental comparison approach but instead propose a memory architecture that represents the set as activations in memory, versus weights in a network.

\citet{mitzenmacher2018model} discusses the benefits and draw-backs of a learned Bloom Filter; distinguishing the empirical false positive rate over the distribution of sets $S$ versus the conditional false positive rate of the model given a particular set $S$. In this paper we focus on the empirical false positive rate because we wish to exploit redundancy in the data and query distribution. \citet{mitzenmacher2018optimizing} also considers an alternate way to combine classical and learned Bloom Filters by `sandwiching' the learned model with pre-filter and post-filter classical Bloom Filters to further reduce space.

\section{Conclusion}
In many situations neural networks are not a suitable replacement to Bloom Filters and their variants. The Bloom Filter is robust to changes in data distribution because it delivers a bounded false positive rate for any sampled subset. However in this paper we consider the questions, ``When might a single-shot neural network provide better compression than a Bloom Filter?''. We see that a model which uses an external memory with an adaptable capacity, avoids BPTT with a feed-forward write scheme, and learns to address its memory, is the most promising option in contrast to popular memory models such as DNCs and LSTMs. We term this model the Neural Bloom Filter due to the analogous incorporation of a hashing scheme, commutative write scheme, and multiplicative read mechanism.

The Neural Bloom Filter relies on settings where we have an off-line dataset (both of stored elements and queries) that we can meta-learn over. In the case of a large database we think this is warranted, a database with $100$K separate set membership data structures will benefit from a single (or periodic) meta-learning training routine that can run on a single machine and sample from the currently stored data, generating a large number of efficient data-structures. We envisage the space cost of the network to be amortized by sharing it across many neural Bloom Filters, and the time-cost of executing the network to be offset by the continuous acceleration of dense linear algebra on modern hardware, and the ability to batch writes and queries efficiently. A promising future direction would be to investigate the feasibility of this approach in a production system.

\subsection*{Acknowledgments}
We thank Peter Dayan, Yori Zwols, Yan Wu, Joel Leibo, Greg Wayne, Andras Gyorgy, Charles Blundell, Daan Weirstra, Pushmeet Kohli, and Tor Lattimor for their insights during this project.

\bibliography{icml2019_conference}
\bibliographystyle{icml2019}

\newpage 

\appendix

\section{Further Model Details}
\label{app:further_model_details}
\subsection{Efficient addressing}
\label{app:efficient_addressing}
We discuss some implementation tricks that could be employed for a production system. 

Firstly the original model description defines the addressing matrix $A$ to be trainable. This ties the number of parameters in the network to the memory size. It may be preferable to train the model at a given memory size and evaluate for larger memory sizes. One way to achieve this is by allowing the addressing matrix $A$ to be non-trainable. We experiment with this, allowing $A \sim \mathcal{N}(\mathbf{0}, \mathbf{I})$ to be a fixed sample of Gaussian random variables. We can think of these as point on a sphere in high dimensional space, the controller network must learn to organize inputs into separate buckets across the surface of the sphere. 

To make the addressing more efficient for larger memory sizes, we experiment with sparsification of the addressing softmax by preserving only the top k components. We denote this sparse softmax $\sigma_k(\cdot )$. When using a sparse address, we find the network can fixate on a subset of rows. This observation is common to prior sparse addressing work \citep{shazeer2017outrageously}. We find sphering the query vector, often dubbed whitening, remedies this (see Appendix \ref{app:sphering} for an ablation). The modified sparse architecture variant is illustrated in Algorithm \ref{alg:nbf_sparse}.

\begin{algorithm}[]
    \caption{Sparse Neural Bloom Filter}
    \label{alg:nbf_sparse}
    \begin{algorithmic}[1] % The number tells where the line numbering should start
    \item[]
    \STATE def sparse\_controller(x):
    \STATE $\quad z \leftarrow f_{enc}(x)$ \\
    \STATE $\quad s \leftarrow f_q(z)$ \COMMENT{Raw query word} \\
    \STATE $\quad q \leftarrow moving\_zca(q)$ \COMMENT{Spherical query} \\
    \STATE $\quad a \leftarrow \sigma_k(q^T A)$ \COMMENT{Sparse address} \\
    \STATE $\quad w \leftarrow f_{w}(z) $  \\ 
    \item[]
    \STATE def sparse\_write(x):
    \STATE $\quad a, w \leftarrow \hbox{sparse\_controller}(x)$
    \STATE $\quad M_{t + 1}[a_{idx}] \leftarrow M_t[a_{idx}] + w a_{val}^T$ \\
    \item[]
    \STATE def sparse\_read(x):
    \STATE $\quad a, w, z \leftarrow \hbox{sparse\_controller}(x)$
    \STATE $\quad r \leftarrow M[a_{idx}] \odot a_{val}$ \\
    \STATE $\quad o \leftarrow f_{out}([r, w, z])$ \\
    \end{algorithmic}
\end{algorithm}

One can avoid the linear-time distance computation $q^TA$ in the addressing operation $\sigma_k(q^TA)$ by using an approximate k-nearest neighbour index, such as locality-sensitive hashing \citep{datar2004locality}, to extract the nearest neighbours from $A$ in $\mathcal{O}(\log m)$ time. The use of an approximate nearest neighbour index has been empirically considered for scaling memory-augmented neural networks \citep{rae2016scaling, kaiser2017learning} however this was used for attention on $M$ directly. As $M$ is dynamic the knn requires frequent re-building as memories are stored or modified. This architecture is simpler --- $A$ is fixed and so the approximate knn can be built once.  

To ensure the serialized size of the network (which can be shared across many memory instantiations) is independent of the number of slots in memory $m$ we can avoid storing $A$. In the instance that it is not trainable, and is simply a fixed sample of random variables that are generated from a deterministic random number generator --- we can instead store a set of integer seeds that can be used to re-generate the rows of $A$. We can let the $i$-th seed $c_i$, say represented as a 16-bit integer, correspond to the set of $16$ rows with indices $16 i, 16i + 1, \ldots, 16i + 15$. If these rows need to be accessed, they can be regenerated on-the-fly by $c_i$. The total memory cost of $A$ is thus $m$ bits, where $m$ is the number of memory slots\footnote{One can replace $16$ with $32$ if there are more than one million slots}.

Putting these two together it is possible to query and write to a Neural Bloom Filter with $m$ memory slots in $\mathcal{O}(\log m)$ time, where the network consumes $\mathcal{O}(1)$ space. It is worth noting, however, the Neural Bloom Filter's memory is often much smaller than the corresponding classical Bloom Filter's memory, and in many of our experiments is even smaller than the number of unique elements to store. Thus dense matrix multiplication can still be preferable - especially due to its acceleration on GPUs and TPUs \citep{jouppi2017datacenter} - and a dense representation of $A$ is not inhibitory. As model optimization can become application-specific, we do not focus on these implementation details and use the model in its simplest setting with dense matrix operations.

\subsection{Moving ZCA}
The moving ZCA was computed by taking moving averages of the first and second moment, calculating the ZCA matrix and updating a moving average projection matrix $\theta_{zca}$. This is only done during training, at evaluation time $\theta_{zca}$ is fixed. We describe the update below for completeness.
\begin{align}
     & \hbox{Input: } s \leftarrow f_{q}(z) \\
     & \mu_{t+1} \leftarrow \gamma \mu_t + (1 - \gamma) \bar{s}  & \hbox{1st moment EMA} \\
    & \Sigma_{t + 1} \leftarrow \gamma \Sigma_{t} + (1 - \gamma) \; s^T s & \hbox{2nd moment EMA} \\
    & U, s, \_ \leftarrow \mathtt{svd}(\Sigma - \mu^2) & \hbox{Singular values} \\
    & W \leftarrow U U^T / \sqrt(s)  & \hbox{ZCA matrix}\\
    & \theta_{zca} \leftarrow \eta \theta_{zca} + (1 - \eta) W & \hbox{ZCA EMA} \\
    & q \leftarrow s \; \theta_{zca} & \hbox{Projected query}
\end{align}
In practice we do not compute the singular value decomposition at each time step to save computational resources, but instead calculate it and update $\theta$ every $T$ steps. We scale the discount in this case $\eta' = \eta / T$.

\subsection{Relation to uniform hashing}
\label{app:hashing_relation}
We can think of the decorrelation of $s$, along with the sparse content-based attention with $A$, as a hash function that maps $s$ to several indices in $M$. For moderate dimension sizes of $s$ ($256$, say) we note that the Gaussian samples in $A$ lie close to the surface of a sphere, uniformly scattered across it. If $q$, the decorrelated query, were to be Gaussian then the marginal distribution of nearest neighbours rows in $A$ will be uniform. If we chose the number of nearest neighbours $k = 1$ then this implies the slots in $M$ are selected independently with uniform probability. This is the exact hash function specification that Bloom Filters assume. Instead we use a continuous (as we choose $k > 1$) approximation (as we decorrelate $s \rightarrow q$ vs Gaussianize) to this uniform hashing scheme, so it is differentiable and the network can learn to shape query representations.

\begin{figure*}[h!]
    \centering
    \includegraphics[width=0.95\textwidth]{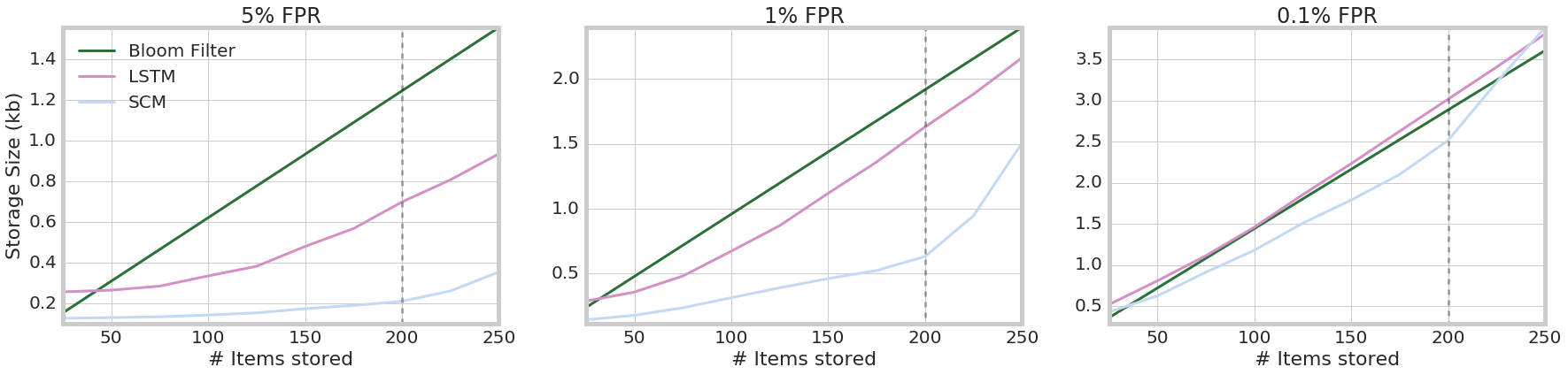}
    \caption{\textbf{Database extrapolation task}. Models are trained up to sets of size $200$ (dashed line). We see extrapolation to larger set sizes on test set, but performance degrades. Neural architectures perform best for larger allowed false positive rates.}
    \label{fig:database_results}
\end{figure*}
\section{Space Comparison}
\label{app:space_comparison}
For each task we compare the model's memory size, in bits, at a given false positive rate --- usually chosen to be $1\%$. For our neural networks which output a probability $p = f(x)$ one could select an operating point $\tau_{\epsilon}$ such that the false positive rate is $\epsilon$. In all of our experiments the neural network outputs a memory (state) $s$ which characterizes the storage set. Let us say \texttt{SPACE(f, $\epsilon$)} is the minimum size of $s$, in bits, for the network to achieve an average false positive rate of $\epsilon$. We could compare \texttt{SPACE(f,$\epsilon$)} with \texttt{SPACE(Bloom Filter,$\epsilon$)} directly, but this would not be a fair comparison as our network $f$ can emit false negatives.

To remedy this, we employ the same scheme as \citet{kraska2018case} where we use a `backup' Bloom Filter with false positive rate $\delta$ to store all false negatives. When $f(x) < \tau_{\epsilon}$ we query the backup Bloom Filter. Because the overall false positive rate is $\epsilon + (1 - \epsilon) \delta$, to achieve a false positive rate of at most $\alpha$ (say $1\%$) we can set $\epsilon = \delta = \alpha / 2$. The number of elements stored in the backup bloom filter is equal to the number of false negatives, denoted $n_{fn}$. Thus the total space can be calculated, \texttt{TOTAL\_SPACE(f,$\alpha$) = SPACE(f,$\frac{\alpha}{2}$) +  $n_{fn}$ * SPACE(Bloom Filter,$\frac{\alpha}{2}$)}. We compare this quantity for different storage set sizes.

\section{Model Size}
\label{app:encoder}
For the MNIST experiments we used a $3$-layer convolutional neural network with $64$ filters followed by a two-layer feed-forward network with $64 \& 128$ hidden-layers respectively. The number of trainable parameters in the Neural Bloom Filter (including the encoder) is $243,437$ which amounts to $7.8$Mb at $32$-bit precision. We did not optimize the encoder architecture to be lean, as we consider it part of the library in a sense. For example, we do not count the size of the hashing library that an implemented Bloom Filter relies on, which may have a chain of dependencies, or the package size of TensorFlow used for our experiments. Nevertheless we can reason that when the Neural Bloom Filter is $4$kb smaller than the classical, such as for the non-uniform instance-based familiarity in Figure \ref{fig:mnist_tasks}b, we would expect to see a net gain if we have a collection of at least $1,950$ data-structures. We imagine this could be optimized quite significantly, by using $16$-bit precision and perhaps using more convolution layers or smaller feed-forward linear operations.

For the database experiments we used an LSTM character encoder with $256$ hidden units followed by another $256$ feed-forward layer. The number of trainable parameters in the Neural Bloom Filter $419,339$ which amounts to $13$Mb. One could imagine optimizing this by switching to a GRU or investigating temporal convolutions as encoders. 

\section{Hyper-Parameters}
\label{app:hyperparams}
We swept over the following hyper-parameters, over the range of memory sizes displayed for each task. We computed the best model parameters by selecting those which resulted in a model consuming the least space as defined in Appendix \ref{app:space_comparison}. This depends on model performance as well as state size. The Memory Networks memory size was fixed to equal the input size (as the model does not arbitrate what inputs to avoid writing).
\begin{table}[h]
    \centering
    \begin{tabular}{lr}
        Memory Size {\small (DNC, NBF)} & \{2, 4, 8, 16, 32, 64\} \\
        Word Size {\small (MemNets, DNC, NBF)} & \{2, 4, 6, 8, 10\} \\
        Hidden Size {\small (LSTM)} & \{2, 4, 8, 16, 32, 64\} \\
        Sphering Decay $\eta$ {\small (NBF)}  & \{0.9, 0.95, 0.99\} \\
        Learning Rate {\small (all)} & \{1e-4, 5e-5\} \\
    \end{tabular}
    \caption{Hyper-parameters considered}
    \label{tab:hyperparams}
\end{table}
\section{Experiment Details}
\label{app:sampling}
For the class-based familiarity task, and uniform sampling task, the model was trained on the training set and evaluated on the test set. For the class-based task sampling, a class is sampled at random and $S$ is formed from a random subset of images from that class. The queries $q$ are chosen uniformly from either $S$ or from images of a different class.

For the non-uniform instance-based familiarity task we sampled images from an exponential distribution. Specifically we used a fix permutation of the training images, and from that ordering chose $p(i_{th} \hbox{ image}) \propto 0.999^i$ for the images to store. The query images were selected uniformly. We used a fixed permutation (or shuffle) of the images to ensure most probability mass was not placed on images of a certain class. I.e. by the natural ordering of the dataset we would have otherwise almost always sampled $0$ images. This would be confounding task non-uniformity for other latent structure to the sets. Because the network needed to relate the image to its frequency of occurence for task, the models were evaluated on the training set. This is reasonable as we are not wishing for the model to visually generalize to unseen elements in the setting of this exact-familiarity task. We specifically want the network weights to compress a map of image to probability of storage.

For the database task a universe of $2.5M$ unique tokens were extracted from GigaWord v5. We shuffled the tokens and placed $2.3$M in a training set and $250$K in a test set. These sets were then sorted alphabetically. A random subset, representing an SSTable, was sampled by choosing a random start index and selecting the next $n$ elements, which form our set $S$. Queries are sampled uniformly at random from the universe set. Models are trained on the training set and evaluated on the test set.

\section{Database Extrapolation Task}
\label{app:database_extrapolation}
We investigate whether neural models are able to extrapolate to larger test sizes. Using the database task setup, where each set contains a contiguous set of sorted strings; we train both the Neural Bloom Filter and LSTM on sets of sizes 2 - 200. We then evaluate on sets up to 250, i.e. a 25\% increase over what is observed during training. This is to emulate the scenario that we train on a selection of databse tablets, but during evaluation we may observe some tablets that are slightly larger than those in the training set. 
Both the LSTM and Neural Bloom Filter are able to solve the task, with the Neural Bloom Filter using significantly less space for the larger allowed false positive rate of 5\% and 1\%. We do see the models' error increase as it surpasses the maximum training set size, however it is not catastrophic. Another interesting trend is noticeable; the neural models have higher utility for larger allowed false positive rates. This may be because of the difficulty in training the models to an extremely low accuracy.

\section{Effect of Sphering}
\label{app:sphering}
\begin{figure}[h]
    \centering
    \includegraphics[width=0.8\columnwidth]{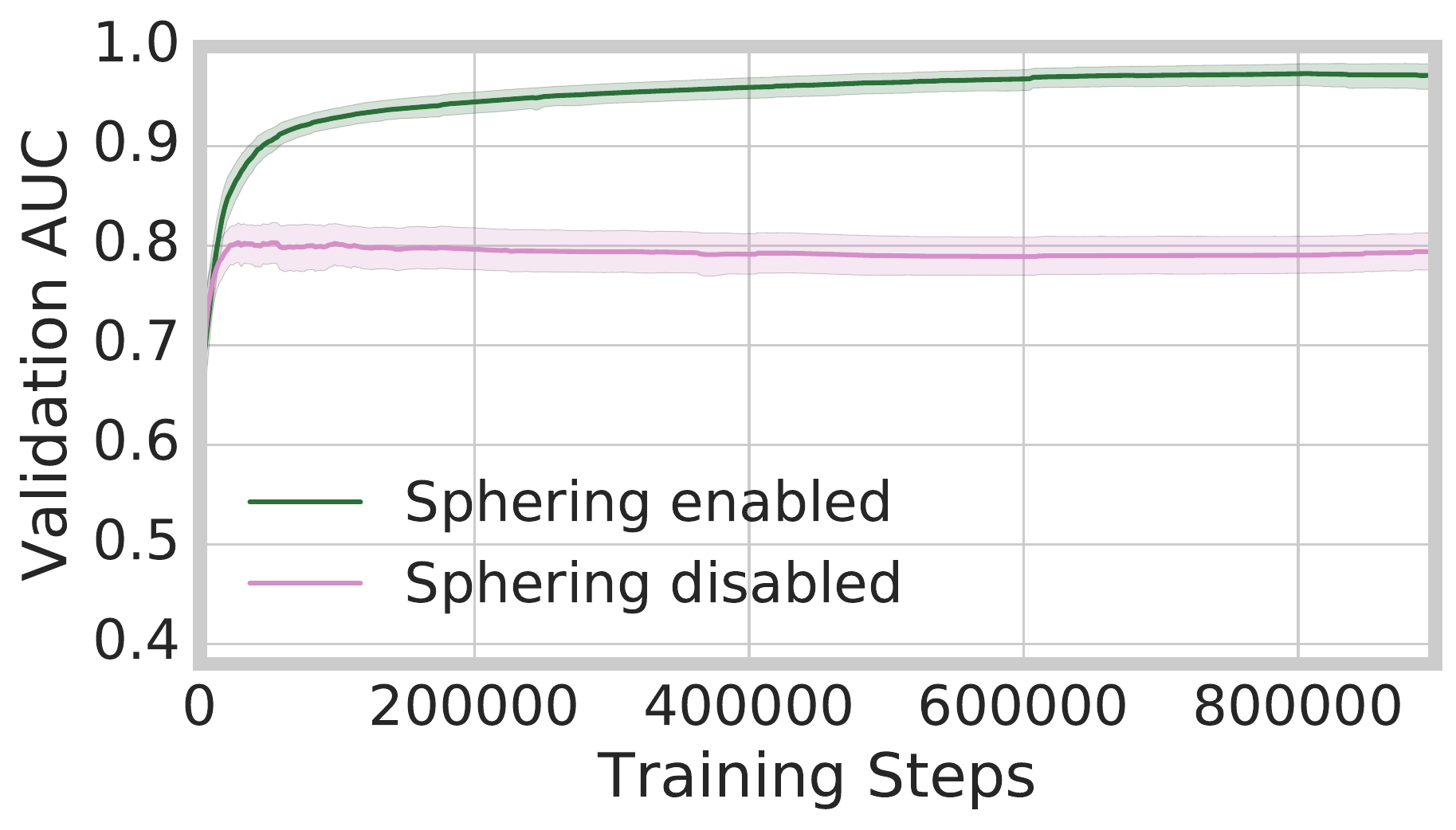}
    \caption{For sparse addresses, sphering enables the model to learn the task of set membership to high accuracy.}
    \label{fig:spherize_auc}
\end{figure}

\begin{figure}[h]
    \centering
    \includegraphics[width=0.8\columnwidth]{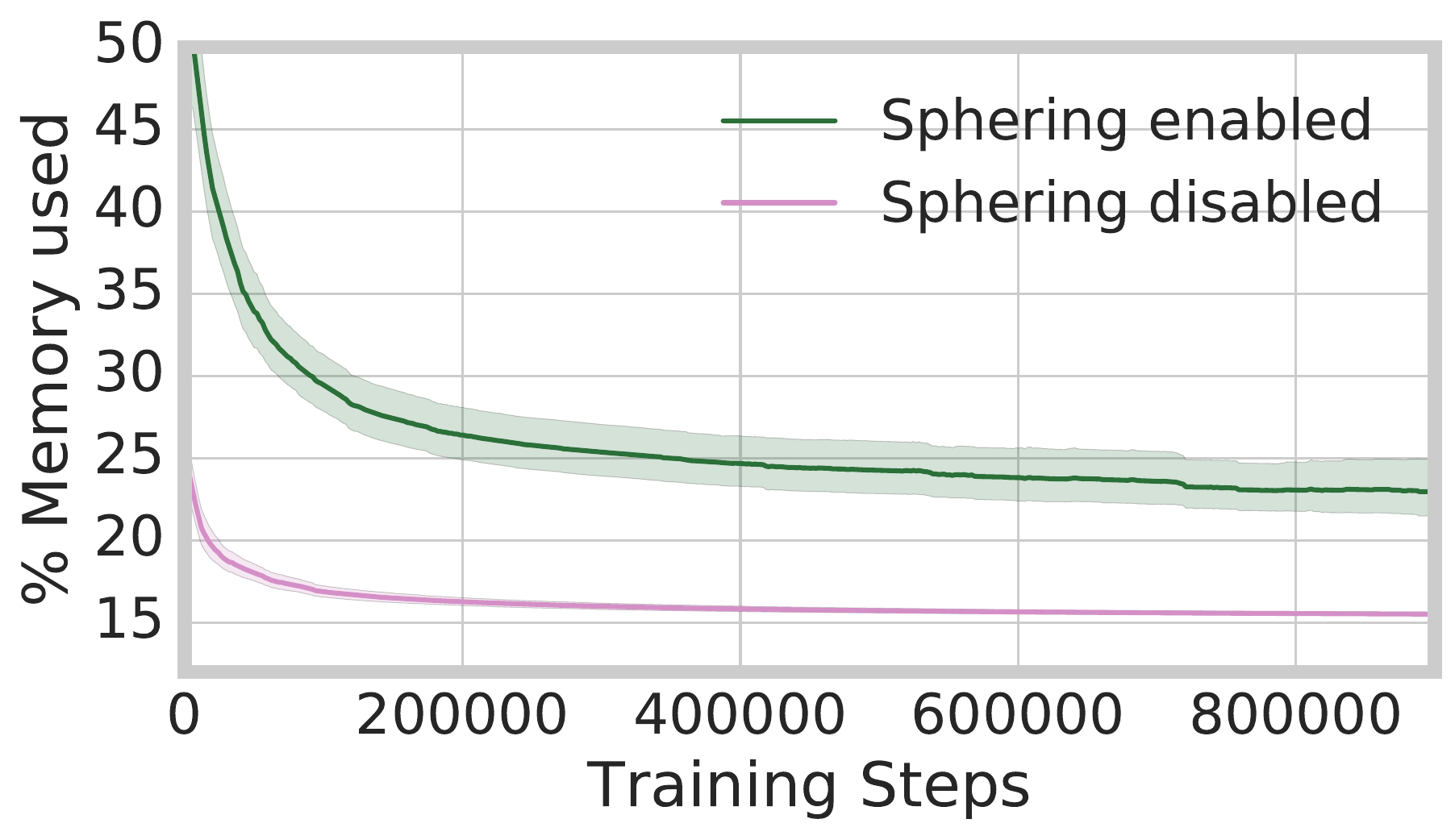}
    \caption{For sparse addresses, sphering the query vector leads to fewer collisions across memory slots and thus a higher utilization of memory.}
    \label{fig:spherize_memory_fill}
\end{figure}

We see the benefit of sphering in Figure \ref{fig:spherize_auc} where the converged validation performance ends up at a higher state. Investigating the proportion of memory filled after all elements have been written in Figure \ref{fig:spherize_memory_fill}, we see the model uses quite a small proportion of its memory slots. This is likely due to the network fixating on rows it has accessed with sparse addressing, and ignoring rows it has otherwise never touched --- a phenomena noted in \citet{shazeer2017outrageously}. The model finds a local minima in continually storing and accessing the same rows in memory. The effect of sphering is that the query now appears to be Gaussian (up to the first two moments) and so the nearest neighbour in the address matrix A (which is initialized to Gaussian random variables) will be close to uniform. This results in a more uniform memory access (as seen in Figure \ref{fig:spherize_memory_fill}) which significantly aids performance (as seen in Figure \ref{fig:spherize_auc}).


\section{Timing Benchmark}
\label{app:timing}
We use the Neural Bloom Filter network architecture for the large database task (Table \ref{tab:database_large}). The network uses an encoder LSTM with $256$ hidden units over the characters, and feeds this through a $256$ fully connected layer to encode the input. A two-layer $256$-hidden-unit MLP is used as the query architecture. The memory and word size is $8$ and $4$ respectively, and so the majority of the compute is spent in the encoder and query network. We compare this with an LSTM containing $32$ hidden units. We benchmark the single-query latency of the network alongside the throughput of a batch of queries, and a batch of inserts. The Neural Bloom Filter and LSTM is implemented in TensorFlow without any custom kernels or specialized code. We benchmark it on the CPU (Intel(R) Xeon(R) CPU E5-1650 v2 @ 3.50GHz) and a GPU (NVIDIA Quadro P6000). We compare to empirical timing results published in a query-optimized Bloom Filter variant \citep{chen2007new}.

It is worth noting, in several Bloom Filter applications, the actual query latency is not in the critical path of computation. For example, for a distributed database, the network latency and disk access latency for one tablet can be orders of magnitude greater than the in-memory latency of a Bloom Filter query. For this reason, we have not made run-time a point of focus in this study, and it is implicitly assumed that the neural network is trading off greater latency for less space. However it is worth checking whether run-time could be prohibitive.

\end{document}